\documentclass[10pt,twocolumn,letterpaper]{article}

\usepackage[pagenumbers]{cvpr} 

\definecolor{cvprblue}{rgb}{0.21,0.49,0.74}
\usepackage[pagebackref,breaklinks,colorlinks,allcolors=cvprblue]{hyperref}

\def\paperID{3325} 
\def\confName{CVPR}
\def\confYear{2025}

\RequirePackage{amsmath}
\RequirePackage[dvipsnames]{xcolor}
\RequirePackage{subcaption}
\RequirePackage{todonotes}
\RequirePackage{multirow}
\RequirePackage{colortbl}
\RequirePackage{xcolor}
\RequirePackage{array}
\definecolor{colorTrd}{rgb}{0.95,0.95,0.65}
\definecolor{colorSnd}{rgb}{1, 0.85, 0.7}
\definecolor{colorFst}{rgb}{1, 0.7, 0.7}

\title{Detail-Preserving Latent Diffusion for Stable Shadow Removal}

\author{Jiamin Xu\\
Hangzhou Dianzi University\\
{\tt\small superxjm@yeah.net}
\and
Yuxin Zheng\\
Hangzhou Dianzi University\\
{\tt\small yuxin6@hdu.edu.cn}
\and
Zelong Li\\
Hangzhou Dianzi University\\
{\tt\small jokerli@hdu.edu.cn}
\and
Chi Wang\\
Zhejiang University\\
{\tt\small wangchi1995@zju.edu.cn}
\and
Renshu Gu,\\
Hangzhou Dianzi University\\
{\tt\small renshugu@hdu.edu.cn}
\and
Weiwei Xu\\
Zhejiang University\\
{\tt\small xww@cad.zju.edu.cn}
\and
Gang Xu\\
Hangzhou Dianzi University\\
{\tt\small gxu@hdu.edu.cn}
}

\begin{document}
\maketitle
\begin{abstract}

Achieving high-quality shadow removal with strong generalizability is challenging in scenes with complex global illumination. Due to the limited diversity in shadow removal datasets, current methods are prone to overfitting training data, often leading to reduced performance on unseen cases. To address this, we leverage the rich visual priors of a pre-trained Stable Diffusion (SD) model and propose a two-stage fine-tuning pipeline to adapt the SD model for stable and efficient shadow removal. In the first stage, we fix the VAE and fine-tune the denoiser in latent space, which yields substantial shadow removal but may lose some high-frequency details. To resolve this, we introduce a second stage, called the detail injection stage. This stage selectively extracts features from the VAE encoder to modulate the decoder, injecting fine details into the final results. Experimental results show that our method outperforms state-of-the-art shadow removal techniques. The cross-dataset evaluation further demonstrates that our method generalizes effectively to unseen data, enhancing the applicability of shadow removal methods.

\end{abstract}
\section{Introduction}
\label{sec:intro}

Shadows are natural phenomena that occur when an obstacle blocks global light transport. They provide crucial cues for interpreting the three-dimensional structure of objects and scenes~\cite{zhang1999shape}. However, shadows can also hinder computer vision tasks, such as object segmentation~\cite{he2017mask}, tracking~\cite{sanin2010improved}, and intrinsic decomposition~\cite{li2018learning, li2022physically, nestmeyer2020learning, ye2023intrinsicnerf}. In recent years, deep-learning-based methods for shadow removal~\cite{guo2023shadowformer,xiao2024homoformer,guo2023shadowdiffusion,mei2024latent} have shown significant advancements by learning pixel-wise mappings between shadowed images and their corresponding ground-truth shadow-free versions in a fully supervised manner. However, such approaches often risk overfitting the limited training datasets~\cite{le2019shadow,qu2017deshadownet,vasluianu2023wsrd}, leading to inadequate generalization.

\begin{figure}[t]
\centering
\includegraphics[width=\linewidth]{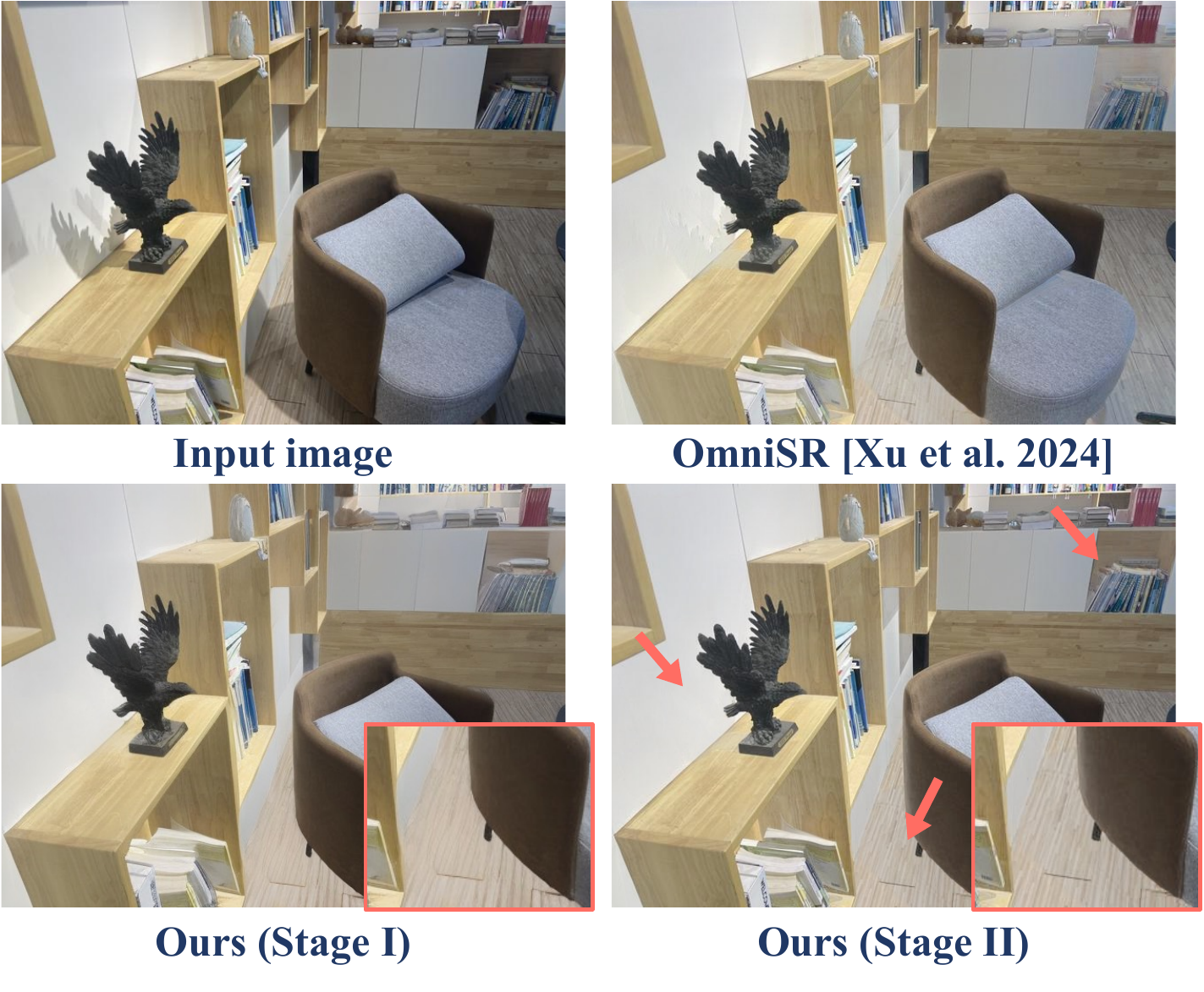}
\caption{\emph{Top}: For complex shadows in indoor scenes, current methods struggle to completely remove shadows, such as those behind the eagle sculpture. \emph{Bottom}: Our method effectively removes the shadows, and with our second stage, the details from the original input, such as the wooden texture shown in the cropped image, are preserved.}
\label{fig:teaser} 
\end{figure}

Drawing inspiration from recent advancements in zero-shot dense prediction~\cite{ke2024repurposing,ye2024stablenormal}, we leverage the robust and comprehensive visual priors of the pre-trained Stable Diffusion (SD) model~\cite{rombach2022high}, which has been trained on billions of images, to enhance the generalization capability of shadow removal. However, developing a fine-tuning strategy that effectively removes shadows while preserving clear details in both shadowed and non-shadowed areas is not trivial. Unlike low-frequency depth and normal prediction tasks~\cite{ke2024repurposing,ye2024stablenormal}, applying the SD model to shadow removal can result in detail loss and misalignment between the conditional image and the shadow-free output (Fig.~\ref{fig:teaser}). It is because the SD model, which consists of a Variational Autoencoder (VAE) and a Latent Diffusion Model (LDM), is designed for text-to-image generation and can lead to lossy compression during the mapping from pixel space ($W \times H \times 3$) to latent space ($\frac{W}{8} \times \frac{H}{8} \times 4$).

To achieve stable and detail-preserving shadow removal using the pre-trained SD priors, we introduce a two-stage pipeline: 1) eliminating shadows in latent space by fixing the pre-trained VAE and fine-tuning the LDM and 2) injecting shadow-aware multi-resolution details from the original input into the VAE decoding process.

In the first stage, our method retains the pre-trained VAE and fine-tunes the LDM by conditioning it on the latent features of the input shadow image. We do not fine-tune the VAE, as it effectively handles shadow-free images despite not being specifically trained for this purpose, and the scale of our shadow-free images is limited. Our findings indicate that the latent space from the VAE, combined with the generative priors from the LDM, enhances shadow removal tasks, particularly when annotated training data is scarce. By operating in a low-resolution latent space rather than high-resolution pixel space, our method performs global self-attention with acceptable memory consumption, capturing more global contextual information compared to convolution and window-based self-attention~\cite{liu2021swin}. 

The fine-tuned SD model in the first stage produces visually appealing results, but it may lose some local details. To further enhance the quality of shadow removal, we propose a second stage for injecting shadow-free details. In this stage, we introduce a Detail Injection~(DI) module that enhances the decoded latent features from the pre-trained VAE decoder by incorporating fine details from the input images. The VAE parameters are kept fixed. To prevent shadows from being reintroduced as ``details'', our DI module is equipped with shadow perception capabilities. Inspired by the observation that the shadow mask in the current shadow removal dataset is derived from the binary difference between shadowed and shadow-free images, we utilize concatenated features from the input shadow images and the coarse, shadow-free images decoded from the LDM output as cues for shadow-free detail injection. By combining features from both image types, the DI module can implicitly detect shadows and learn to inject shadow-free details into the VAE decoders.

Note that Refusion~\cite{luo2023refusion} also performs latent-space diffusion for shadow removal. It captures input image information in the decoder~\cite{luo2023refusion} using a U-Net architecture specifically designed for large image shadow removal. However, the primary difference lies in the network design and fine-tuning strategy. Our method is designed to leverage the high-quality priors of the pre-trained VAE and LDM in the SD model, resulting in superior shadow removal quality. Our method can also handle high-resolution images (e.g., 1920×1440 in the WRSD dataset~\cite{vasluianu2023wsrd}), providing greater shadow removal efficiency compared to patch-based diffusion methods~\cite{luo2024diff}.

Overall, our contributions can be summarized as follows.
\begin{itemize}

\item We propose a two-stage fine-tuning pipeline to transform a pre-trained Stable Diffusion model into an image-conditional shadow-free image generator. This approach enables robust, high-resolution shadow removal without an input shadow mask.

\item We introduce a shadow-aware detail injection module that utilizes the VAE encoder features to modulate the pre-trained VAE decoders, selectively aligning per-pixel details from the input image with those in the output shadow-free image.

\item Extensive experimental results on public datasets demonstrate that the proposed method significantly outperforms state-of-the-art shadow removal methods. Additionally, we show that our method exhibits better generalizability by training on one dataset and testing on another.

\end{itemize}









\section{Related Work}
\label{sec:related}


\subsection{Deep Shadow Removal}

\begin{figure*}[t!]
\centering
\includegraphics[width=1.0\linewidth]{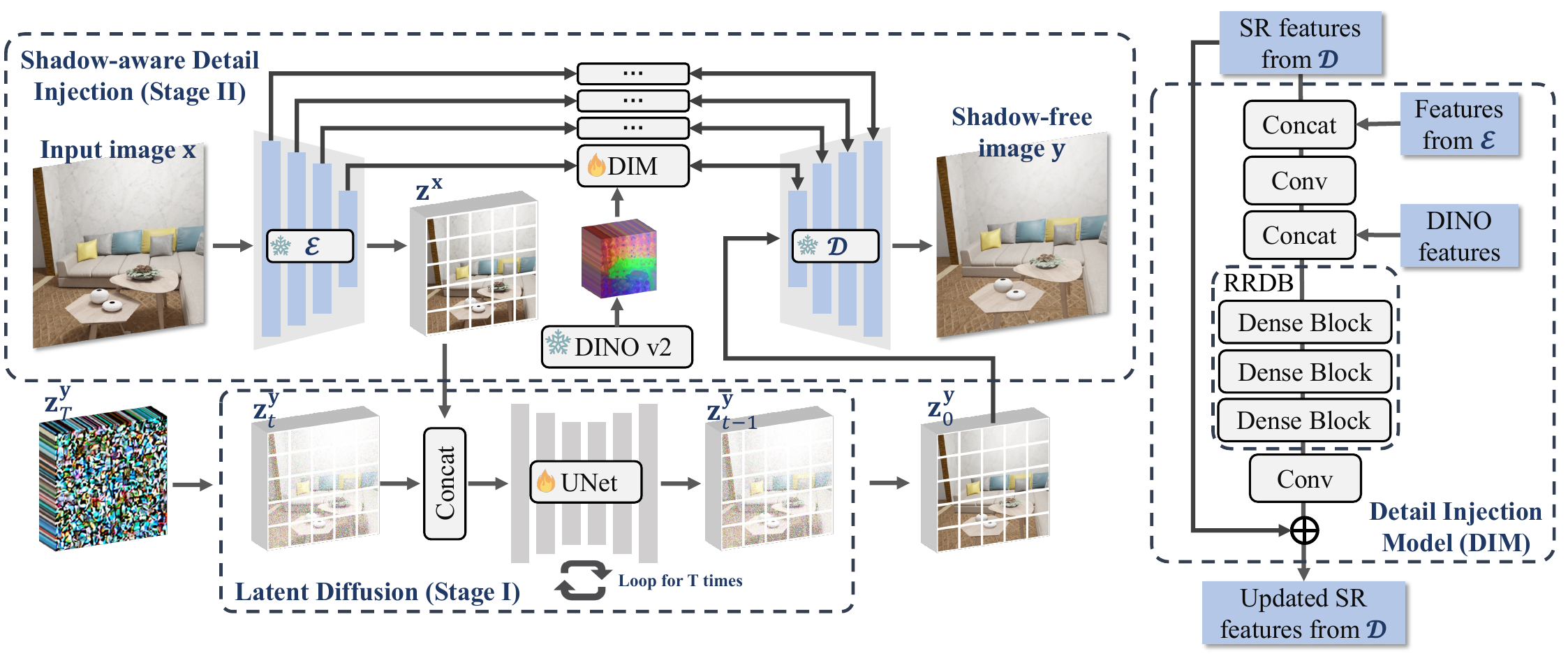}
\caption{\textbf{Our proposed network.} We propose a two-stage shadow removal network based on Stable Diffusion (SD). (1) In the first stage, as shown in the bottom half, we fine-tune the pre-trained UNet in SD within the latent space defined by SD's pre-trained VAE ($\mathcal{E}$ and $\mathcal{D}$). We found that the pre-trained latent space can effectively represent shadow-free images. (2) In the second stage, as shown in the top half, we modulate the VAE decoder $\mathcal{D}$ by selectively adding features from the VAE encoder $\mathcal{E}$ using a Detail Injection Model (DIM). The model consists of multiple RRDB layers, which inject shadow-free texture details into the decoder features. With these two stages, our proposed network can generate high-quality, shadow-free images that preserve fine details.}
\label{fig:pipeline} 
\end{figure*}

Deep learning methods have significantly advanced shadow removal, enabling end-to-end learning of a mapping from input images to corresponding shadow-free images, with or without the aid of shadow masks. For example, DeshadowNet~\cite{qu2017deshadownet} combines multi-level features to predict a shadow matte, thereby enhancing the efficiency of shadow removal. Hu et al.~\cite{cun2020towards,hu2019mask} utilize direction-aware spatial context for precise shadow detection and elimination. Fu et al.~\cite{fu2021auto} formulate the shadow removal task as a multiple exposure image fusion problem. BMNet~\cite{zhu2022bijective} introduces a bijective mapping that integrates shadow removal with shadow generation, improving overall performance in shadow removal tasks. 

More recently, ShadowFormer~\cite{guo2023shadowformer} presents a Swin-transformer-based shadow removal network that harnesses multi-scale attention to exploit contextual relationships between shadowed and non-shadowed regions. Meanwhile, DMTN~\cite{liu2023decoupled} proposes a decoupled multi-task network that learns decomposed features specifically tailored for shadow removal. ShadowRefiner~\cite{dong2024shadowrefiner} introduces a mask-free model that integrates spatial and frequency domain representations for image shadow removal, achieving top results in the Perceptual Track of the NTIRE 2024 Image Shadow Removal Challenge~\cite{vasluianu2024ntire}. Xu et al.~\cite{xu2024omnisr} propose a path-tracing pipeline to generate shadow removal data that considers indirect illumination in indoor scenes. They also introduce a semantics and geometry-aware network for efficient mask-free shadow removal. HomoFormer~\cite{xiao2024homoformer} designs random shuffle and inverse shuffle operations to create a homogenized appearance of shadows, allowing for effective processing by local self-attention layers.

\subsection{Diffusion-based Shadow Removal}

Diffusion models~\cite{ho2020denoising,song2021scorebased} are generative models that learn data distributions through Gaussian noise blurring and reverse denoising. Initially developed for image generation, these models have become prominent in applications including image super-resolution~\cite{saharia2022image,wang2024exploiting}, depth/normal prediction~\cite{ke2024repurposing,ye2024stablenormal}, and shadow removal~\cite{luo2023refusion}. In the context of shadow removal, ShadowDiffusion~\cite{guo2023shadowdiffusion} introduces a diffusion model that generates both a shadow-free image and a refined shadow mask, using shadow degradation as a prior and employing an unrolling diffusion process to eliminate shadows. DeS3~\cite{jin2024des3} enhances ShadowDiffusion by incorporating an adaptive attention mechanism and ViT similarity loss, removing the dependence on shadow masks. Guo et al.\cite{guo2023boundary} propose a diffusion-based unsupervised shadow removal pipeline that utilizes a boundary-aware divide and conquer strategy. Diff-Shadow\cite{luo2024diff} employs a globally guided diffusion model pipeline that focuses on the patch and global information, enabling shadow removal in images of arbitrary resolution. Liu et al.~\cite{liu2024recasting} first separate reflectance and illumination, then use a diffusion model to correct degraded lighting in shadowed areas, followed by the progressive restoration of texture details through an illumination-guided texture restoration module. 

There are also methods that utilize a latent space to enhance diffusion-based shadow removal. Refusion~\cite{guo2023boundary} presents a U-Net-based latent diffusion model that utilizes mean-reverting stochastic differential equations for high-resolution image restoration tasks. The model performs image diffusion in a low-resolution latent space while preserving details through hidden-connections. However, being end-to-end trained on the shadow-removal dataset, it suffers from limited generalizability. Mei et al.~\cite{mei2024latent} conduct diffusion-based shadow removal with a shadow mask at both the general and instance levels. They improve the diffusion process by conditioning on a learned latent feature space that captures the characteristics of shadow-free images and propose fusing noise features with the diffusion network to alleviate potential local optima.

Unlike existing diffusion-based methods, we introduce a two-stage fine-tuning strategy that fully utilizes the visual priors from the pre-trained Stable Diffusion (SD) model~\cite{rombach2022high}. With our latent shadow removal and shadow-aware detail injection stages, our method demonstrates strong generalization capability while effectively preserving non-shadow details.

\section{Proposed Method}
\label{sec:method}

\subsection{Overview} 

Given a dataset of shadow and shadow-free image pairs $\left<\mathbf{x}, \mathbf{y}\right> \in \mathcal{X}$, our goal is to train a network that can: 1. thoroughly eliminate shadows from the input image, and 2. preserve the content of all non-shadow areas. Although the first requirement focuses on the shadow (umbra) regions, the second is more concerned with non-shadow areas. These two objectives can be contradictory, making it challenging to design a single network that effectively addresses both. Therefore, we propose a two-stage shadow removal framework to robustly eliminate various types of shadows across different scenarios (Fig.~\ref{fig:pipeline}).

In the \textbf{first stage}, as shown at the bottom of Fig.~\ref{fig:pipeline}, we perform shadow removal in latent space (Sec.~\ref{sec:sr_in_latent}). We condition the Stable Diffusion (SD) model~\cite{rombach2022high} using the latent code of the input shadow image $\mathbf{z}^\mathbf{x}$ and fine-tune a pretrained SD model to generate the latent code of a shadow-free image $\mathbf{z}^\mathbf{y}_0$. As shown in Fig.~\ref{fig:teaser}, this latent code can be decoded into a coarse image that effectively eliminates shadows, although some details may be lost.

In the \textbf{second stage}, as illustrated at the top of Fig.~\ref{fig:pipeline}, we perform detail injection using the encoder $\mathcal{E}$ and decoder $\mathcal{D}$ within SD (Sec.\ref{sec:SDI}). We fix the pretrained encoder and decoder and employ a set of Detail Injection (DI) models to integrate shadow-free details into the decoder features. These details are derived from each encoder layer. After training on pairs of shadow and shadow-free images, the DI models learn to inject details from the input image that are free of shadows.

\begin{figure}[t!]
\centering
\includegraphics[width=1.0\linewidth]{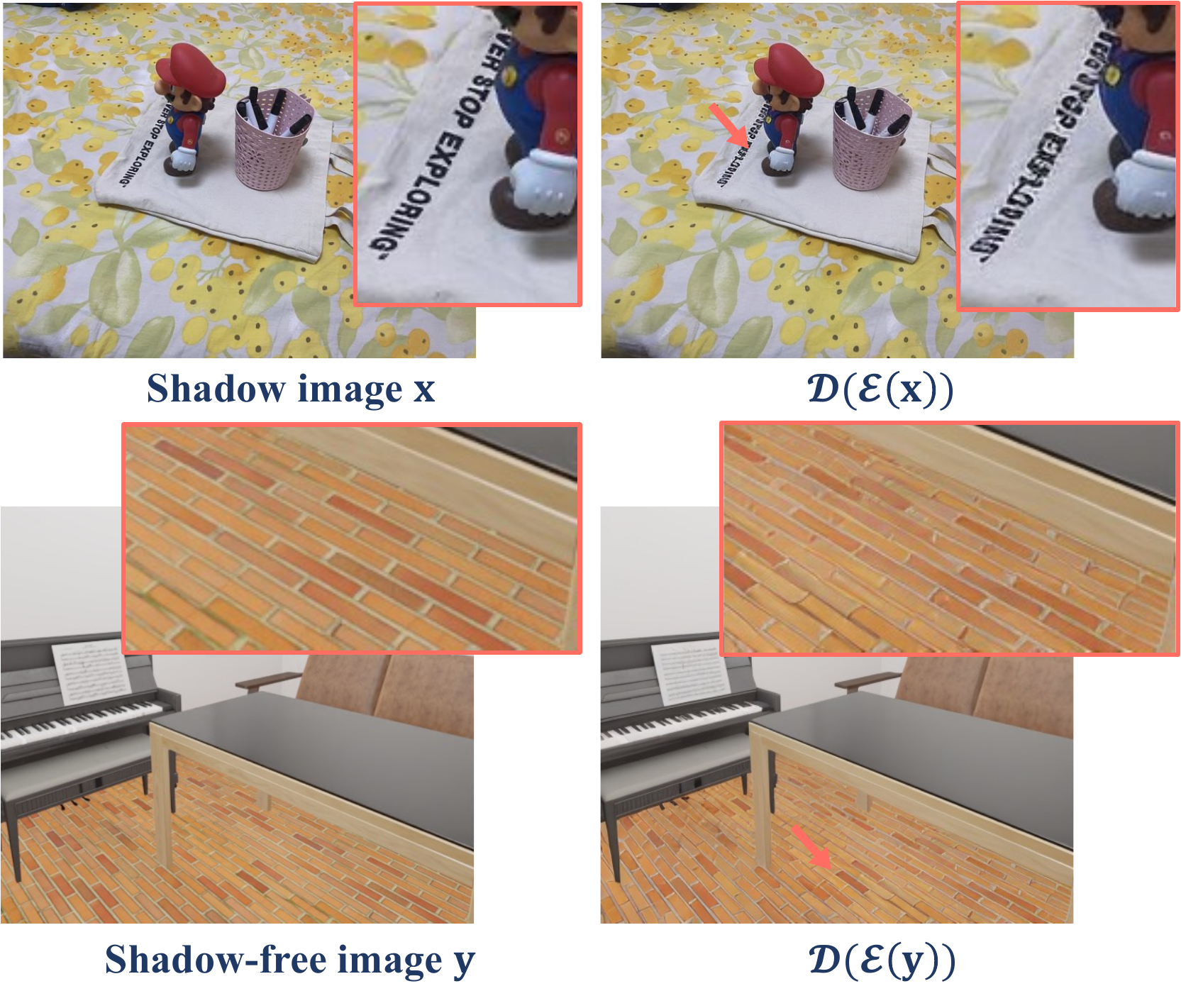}
\caption{\textbf{Latent space in VAE.} We apply the pre-trained encoding and decoding process on a shadow image $\mathbf{x}$ or a shadow-free image 
$\mathbf{y}$. For a shadow-free image, this process does not introduce additional shadows, meaning the pre-trained latent space can also represent the shadow-free image. However, as shown in the zoomed-in view, some details, like text, may be lost in this process.}
\label{fig:vae_result} 
\end{figure}


\subsection{Shadow Removal in Latent Space} 
\label{sec:sr_in_latent}


Building on recent advancements in dense prediction tasks~\cite{ke2024repurposing,ye2024stablenormal,he2024lotus}, we formulate the shadow removal task in the first stage as an image-conditioned annotation generation problem using SD. This approach conducts the diffusion process within a low-dimensional latent space generated by a pair of encoder and decoder $\left<\mathcal{E}, \mathcal{D}\right>$, where $\mathcal{E}(\textbf{x}) = \mathbf{z^x}$ and $\mathcal{D}(\mathbf{z^x}) \approx \textbf{x}$.

Compared to the diffusion model~\cite{ho2020denoising}, Stable Diffusion offers enhanced generation stability and training efficiency. However, given the limited number of shadow and shadow-free image pairs, we opt for a fine-tuning strategy rather than training a SD model from scratch. We directly utilize the pre-trained $\mathcal{E}$ and $\mathcal{D}$ in SD and focus on fine-tuning the denoiser U-Net, as we have found that the pre-trained $\mathcal{E}$ and $\mathcal{D}$ perform well for both shadow and non-shadow images, even though they were not specifically trained on shadow-free images (Fig.~\ref{fig:vae_result}).

\paragraph{Fine-tuning Process.}

The fine-tuning for Stable Diffusion~\cite{ho2020denoising} relies on forward-noising and reverse-denoising processes that operate within the latent space. In the forward process, Gaussian noise is progressively added over time steps $t \in [1, T]$ to the sample $\mathbf{z^y}$, generating a noisy sample $\mathbf{z^y_t}$:
\begin{equation}
\mathbf{z^y_t} = \sqrt{\overline{\alpha}_t}\mathbf{z^y} + \sqrt{1-\overline{\alpha}_t}\mathbf{\epsilon}, 
\end{equation}
where $\epsilon\sim\mathcal{N}(0, I)$, $\overline{\alpha}_t := \prod_{s=1}^t (1 - \beta_s)$, and $\beta_1$, $\beta_2$, $\dots$, $\beta_T$ defines the noise schedule over $T$ steps. In the reverse process, a U-Net~\citep{ronneberger2015u} denoted as $f_\theta$ takes the encoding features of the input image $\mathbf{z^x}$ and the sample $\mathbf{z^y_t}$ as inputs, outputting a prediction of the clean sample $\hat{\mathbf{z}}^\mathbf{y}$:
\begin{equation} \label{eq
} \hat{\mathbf{z}}^\mathbf{y} = f_\theta(\mathbf{z^y_t}, \mathbf{z^x}, t).
\end{equation}

During fine-tuning, we randomly sampling a time step $t \in [1, T]$ and minimizing the corresponding loss function $L_t$:
\begin{equation}  
\mathcal{L}_t = ||\mathbf{z^y}-f_\theta(\mathbf{z^y_t}, \mathbf{z^x}, t)||_2^2, 
\label{eq:z-prediction}
\end{equation}
where $\mathbf{z^y}$ is the encoding features of the ground-truth shadow-free image.

\paragraph{Denoising Process.}

During inference, we use DDIM~\cite{song2020denoising}, which implements an implicit probabilistic model that can significantly reduce the number of denoising steps while maintaining output quality. Formally, the denoising process from $\mathbf{z^y_\tau}$ to $\mathbf{z^y_{\tau-1}}$ is: 
\begin{equation}
\label{eq:ddim}
    \mathbf{z^y_{\tau-1}} = \sqrt{\overline{\alpha}_{\tau-1}} \mathbf{\hat{z}^y_\tau} + \text{direction}(\mathbf{z^y_{\tau}}) + \sigma_\tau\epsilon_\tau, 
\end{equation}
where $\mathbf{\hat{z}^y_\tau}$ is the predicted clean sample at the denoising step $\tau$, $\text{direction}(\mathbf{z^y_{\tau}})$ represents the direction pointing to $\mathbf{z^y_{\tau}}$. $\tau \in \{ \tau_1, \tau_2, \dots, \tau_S \}$ is an increasing sub-sequence of the time-step set $[1,T]$ used for fast sampling.

\begin{figure}[t!]
\centering
\includegraphics[width=1.0\linewidth]{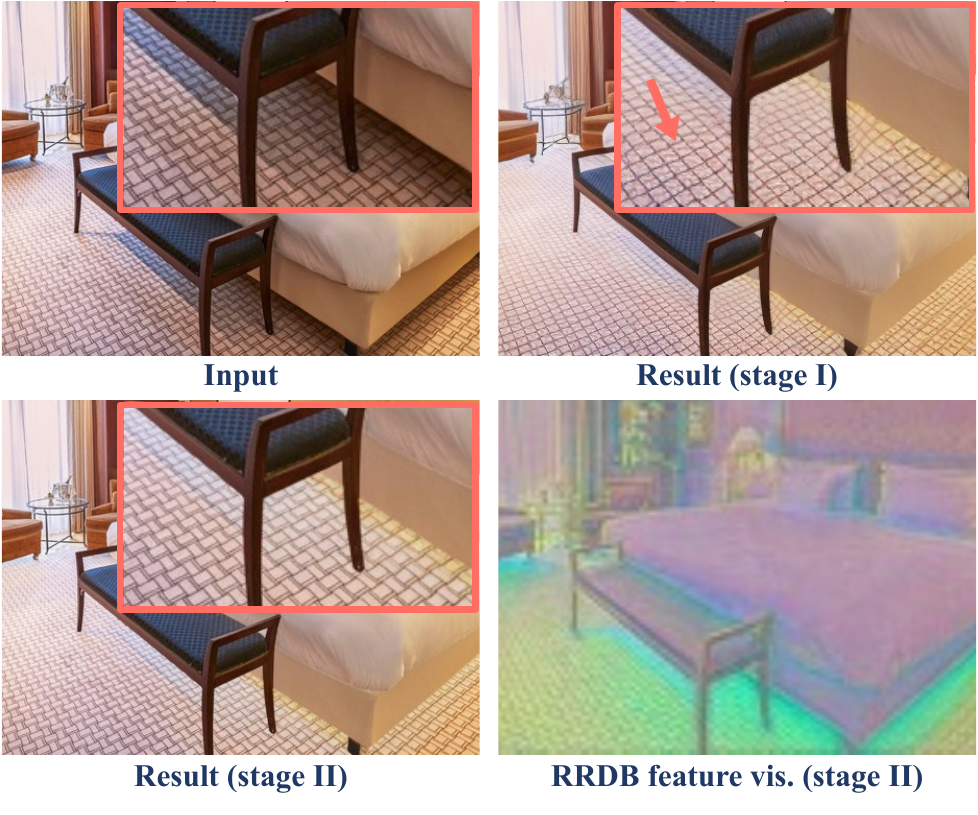}
\caption{\textbf{Some details in the input image, such as the textures on the floor, change during latent space diffusion in stage one.} In stage two, our method preserves these high-quality details while effectively removing shadows. As shown in the bottom-left image, where we map the RRDB features to three dimensions using PCA, we observe that shadow areas have distinct features (in green), indicating that our detail injection model is shadow-aware.}
\label{fig:stage1&2} 
\end{figure}

\subsection{Shadow-aware Detail Injection} 
\label{sec:SDI}

As illustrated in Fig.~\ref{fig:stage1&2}, while the output of the conditional Stable Diffusion is visually striking—especially in its ability to completely remove shadows—it often diverges from the ground truth, particularly in high-detail areas, such as the floor texture. We believe the primary reason for this is the loss of high-frequency information in the latent space. As shown in Fig.~\ref{fig:vae_result}, directly decoding the encoded features $\mathcal{D}(\mathcal{E}(\textbf{x}))$ also results in a loss of detail. 

To eliminate shadows while preserving details, we introduce a shadow-aware detail injection step utilizing a set of DI models. Each model selectively extracts information from a specific intermediate layer of $\mathcal{E}$ to modulate the corresponding intermediate layers of $\mathcal{D}$. Let $\mathbf{e}_i$ and $\mathbf{d}_i$ denote the encoder and decoder features for the $i$-th intermediate layer. The DI model takes the concatenation of $\mathbf{e}_i$ and $\mathbf{d}_i$ as input, output a modulated feature $\Tilde{\mathbf{d}}_i$:
\begin{align}
\mathbf{d}_0&=\mathcal{D}_0(\mathbf{z}^\mathbf{y}_0),\\
\Tilde{\mathbf{d}}_i&=\mathbf{d}_i+\mathrm{Conv}(\mathrm{RRDB}(\mathrm{Conv}(\mathbf{d}_i,\mathbf{e}_{n-i}(\mathbf{z}^\mathbf{x})), f_d(\mathbf{x}))),\notag\\
\mathbf{d}_{i}&=\mathcal{D}_i(\Tilde{\mathbf{d}}_{i-1}),i\in\left[1,3\right],
\hat{\mathbf{y}}=\mathrm{Conv}({\Tilde{\mathbf{d}}_3})\notag
\end{align}
Here, $\mathrm{RRDB}(\cdot)$ refers to the Residual-in-Residual Dense Block module~\cite{zhang2018residual}, which consists of three Dense Blocks. $\hat{\mathbf{y}}$ is prediected shadow-free image. $f_d(\mathbf{x})$ are features extracted using DINO v2~\cite{oquab2023dinov2}. We concatenate the DINO features with $\mathbf{e}_i$ and $\mathbf{d}_i$ after passing through a convolutional layer to enhance the generalizability of the DI models.

Since $\mathbf{e}_i$ and $\mathbf{d}_i$ contain features from the input shadow image $\mathbf{x}$ and the shadow-free latent code
$\mathbf{z}^\mathbf{y}_0$, respectively, our DI models can effectively identify shadow areas by utilizing both features as input. Although the latent codes may lose high-frequency details, they still provide valuable cues for shadow regions, which are primarily low-frequency. As illustrated in Fig.~\ref{fig:stage1&2}, where we visualize the features in the last layer of the RRDB module, we can observe that the shadow areas exhibit distinct feature responses. We present only the RRDB responses for $\mathbf{d}_2$. Please refer to the supplementary for additional results.

\paragraph{Losses.}

To train the DI models, we minimize a total loss $\mathcal{L}_{total}$, which combines an L1 loss and a perceptual loss measured using the Learned Perceptual Image Patch Similarity (LPIPS) metric~\cite{zhang2018perceptual} between the output image $\hat{\mathbf{y}}$ and the ground-truth shadow-free image $\mathbf{y}$:
\begin{align}
\mathcal{L}_{total} &= ||\hat{\mathbf{y}}-\mathbf{y}||_1 + ||\mathrm{LPIPS}(\hat{\mathbf{y}})-\mathrm{LPIPS}(\mathbf{y})||^2_2.
\end{align}

\subsection{Implementation Details.}

\begin{table}[t]
\center
\begin{tabular}{l c c}
\hline
Method & Mean$\uparrow$ & Variance$\downarrow$ \\
\hline
Our stage one ($\epsilon$-pred) & 29.66 & 0.239 \\
Our stage one ($\mathbf{z}_0$-pred) & \textbf{29.95} & \textbf{0.146} \\
\hline
Our stage two & \textbf{35.02} & \textbf{0.160}  \\
DeS3~\cite{jin2024des3} & 31.33 & 1.075  \\
\hline
\end{tabular}
\caption{\textbf{Low variance.} In the first stage, our method achieves reduced variance by using $\mathbf{z}_0$-prediction. In the second stage, our method improves PSNR while maintaining low variance, resulting in robust, high-quality shadow removal results.}
\label{tab:variance}
\end{table}

\begin{table*}[t]
\center
\begin{tabular}{l c c c c c c c c c c}
\hline
\multirow{2}{*}{Method} & \multirow{2}{*}{Year} & \multirow{2}{*}{Mask-free} & \multicolumn{2}{c}{ISTD+ Dataset} & \multicolumn{2}{c}{SRD Dataset} & \multicolumn{2}{c}{INS Dataset} & \multicolumn{2}{c}{WSRD+ Dataset} \\ \cmidrule(lr){4-5}\cmidrule(lr){6-7}\cmidrule(lr){8-9}\cmidrule(lr){10-11}
& & & {PSNR$\uparrow$} & {SSIM$\uparrow$} & {PSNR$\uparrow$} & {SSIM$\uparrow$} & {PSNR$\uparrow$} & {SSIM$\uparrow$} & {PSNR$\uparrow$} & {SSIM$\uparrow$} \\  
\hline
AutoExposure~\cite{fu2021auto} & 2021 & No & 29.45 & 0.861 & 29.24 & 0.938 & 27.91 & 0.957 & 21.66 & 0.752 \\ 
Zhu et al.~\cite{zhu2022efficient}& 2022 & No & --- & --- & 32.05 & 0.965 & --- & --- & --- & --- \\
BMNet~\cite{zhu2022bijective} & 2022 & No & 33.98 & \cellcolor{colorTrd}{0.972} & 31.97 & 0.965 & 27.90 & 0.958 & 24.75 & 0.816 \\ 
ShadowFormer~\cite{guo2023shadowformer} & 2023 & No & \cellcolor{colorSnd}{35.46} & 0.971 & 32.90 & 0.958 & 28.62 & 0.963 & \cellcolor{colorTrd}{25.44} & \cellcolor{colorTrd}{0.820} \\ 
DMTN~\cite{liu2023decoupled}  & 2023 & No & 32.23 & 0.966 & 33.77 & 0.968 & 28.83 & \cellcolor{colorTrd} {0.969} & --- & --- \\
ShadowDiffusion~\cite{guo2023shadowdiffusion} & 2023 & No & 34.63 & 0.967 & \cellcolor{colorFst}{34.73} & \cellcolor{colorSnd}{0.970} & \cellcolor{colorTrd}{29.12} & 0.966 & --- & --- \\ 
HomoFormer~\cite{xiao2024homoformer} & 2024 & No & \cellcolor{colorFst}{35.72} & \cellcolor{colorFst}{0.977} & \cellcolor{colorSnd}{34.36} & \cellcolor{colorFst}{0.977} & 28.98 & 0.965 & --- & --- \\
InstanceShadow~\cite{mei2024latent} & 2024 & No & 34.69 & 0.968 & 31.61 & 0.959 & 28.16 & 0.948 & --- & --- \\ 
\hline
TBRNet~\cite{liu2023shadow} & 2023 & Yes & 31.91 & 0.964 & 31.83 & 0.953 & --- & --- & --- & --- \\ 
Refusion~\cite{luo2023refusion} & 2023 & Yes & 32.41 & 0.961 & 31.60 & 0.949 & 28.13 & 0.958 & 22.32 & 0.738 \\
DeS3~\cite{jin2024des3} & 2024 & Yes & 31.39 & 0.957 & \cellcolor{colorTrd}{34.11} & 0.968 & 27.89 & 0.947 & --- & --- \\
OmniSR~\cite{xu2024omnisr} & 2024 & Yes & 33.34 & 0.970 & 32.87 & \cellcolor{colorTrd}{0.969} & \cellcolor{colorSnd}{30.38} & \cellcolor{colorSnd}{0.973} & \cellcolor{colorSnd}{26.07} & \cellcolor{colorFst}{0.835} \\
Ours & --- & Yes & \cellcolor{colorTrd}{35.19} & \cellcolor{colorSnd}{ 0.974} & 33.63 & 0.968 & \cellcolor{colorFst}{30.56} & \cellcolor{colorFst}{0.975} & \cellcolor{colorFst}{26.26} & \cellcolor{colorSnd}{0.827} \\
\hline
\end{tabular}
\caption{\textbf{Quantitative comparisons on ISTD+, SRD, INS, and WSRD+ datasets.} Best results are highlighted as \colorbox{colorFst}{1st}, \colorbox{colorSnd}{2nd} and \colorbox{colorTrd}{3rd}.}
\label{tab:comparison}
\end{table*}

\paragraph{Reducing variance.}

As shown in Eq.~\ref{eq:z-prediction}, in the LDM fine-tuning stage, we use $\mathbf{z}_0$-prediction rather than the standard $\epsilon$-prediction parameterization for the denoising model. This choice, as recommended by Lotus\cite{he2024lotus}, helps reduce variance during stochastic LDM inference. We compare these two approaches on the ISTD+~\cite{le2019shadow} dataset by running inference on the test set with five different random seeds (1, 2, 3, 4, 5). We calculate the PSNR variance across the five outputs for each image and obtain the average variance. As shown in Table~\ref{tab:variance}, using $\mathbf{z}_0$-prediction in our LDM fine-tuning stage yields higher PSNR and lower variance compared to $\epsilon$-prediction. Additionally, we assess the average variance of our final results after the detail injection stage, and Table~\ref{tab:variance} shows that our final results have significantly lower variance than DeS3~\cite{jin2024des3}, another mask-free shadow removal method based on diffusion models. It indicates that our generative shadow removal method is highly robust to randomness, which is a critical factor for practical use. 






\paragraph{Training details.}

Our model is trained on a GPU server with four GeForce RTX 4090 GPUs using PyTorch 2.0.1~\cite{paszke2017automatic} with CUDA 11.7. We employ the Adam optimizer~\cite{Kingma2015AdamAM} for training. In the first stage, the LDM is fine-tuned for 40 epochs on INS~\cite{xu2024omnisr} and 200 and 300 epochs on ISTD+~\cite{wang2018stacked,le2019shadow} and SRD~\cite{qu2017deshadownet}, respectively. We utilize a constant learning rate of $3 \times 10^{-4}$ with a batch size 16. For the INS dataset, each batch consists of $512 \times 512$ patches, while for the other datasets, batches are composed of randomly cropped $480 \times 480$ patches. 

In the second stage, we maintain the constant learning rate of $5 \times 10^{-4}$ and batch size of 16. The batch shapes for each dataset are consistent with those in the first stage, and the number of epochs for training the detail injection models is 50 for INS and 300 and 200 for ISTD+ and SRD, respectively. For the DINO features~\cite{oquab2023dinov2}, we resize the image to $\frac{14W}{16} \times \frac{14H}{16}$ using bilinear interpolation, extract the DINO features, and apply a $1 \times 1$ convolution layer to obtain a feature map of shape $256 \times \frac{W}{16} \times \frac{H}{16}$. To reduce memory footprint, we only concatenate the DINO features with the first two layers of the VAE decoder. Additional details can be found in the supplementary.
\section{Experiments}



\begin{figure*}[t!]
\centering
\includegraphics[width=1.0\linewidth]{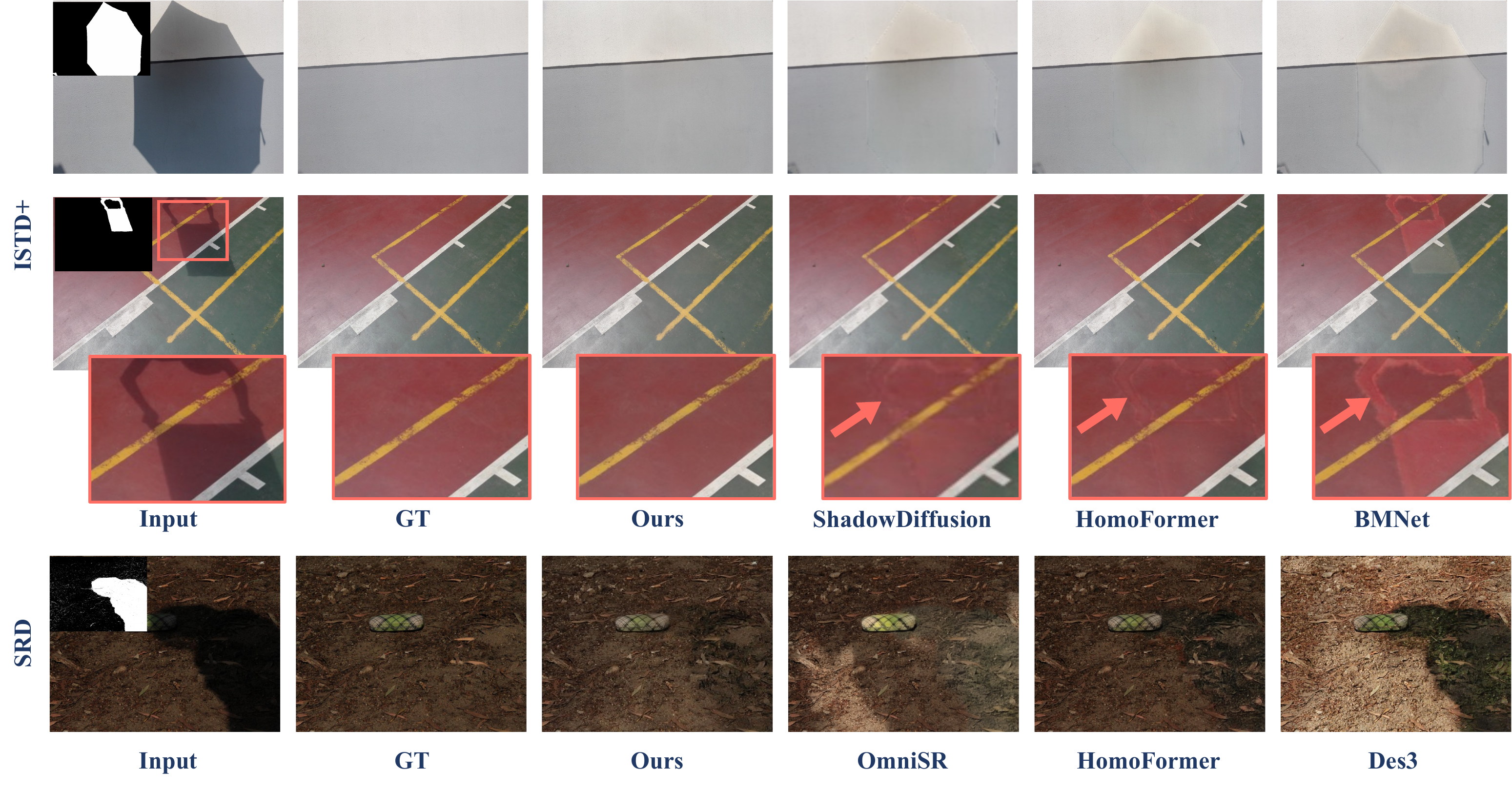}
\caption{\textbf{Comparisons with SOTA shadow removal methods on the ISTD+ and SRD datasets.} The input mask, required by methods that are not mask-free, is shown in the top-left corner.}
\label{fig:comparison} 
\end{figure*}
\begin{figure*}[t!]
\centering
\includegraphics[width=1.0\linewidth]{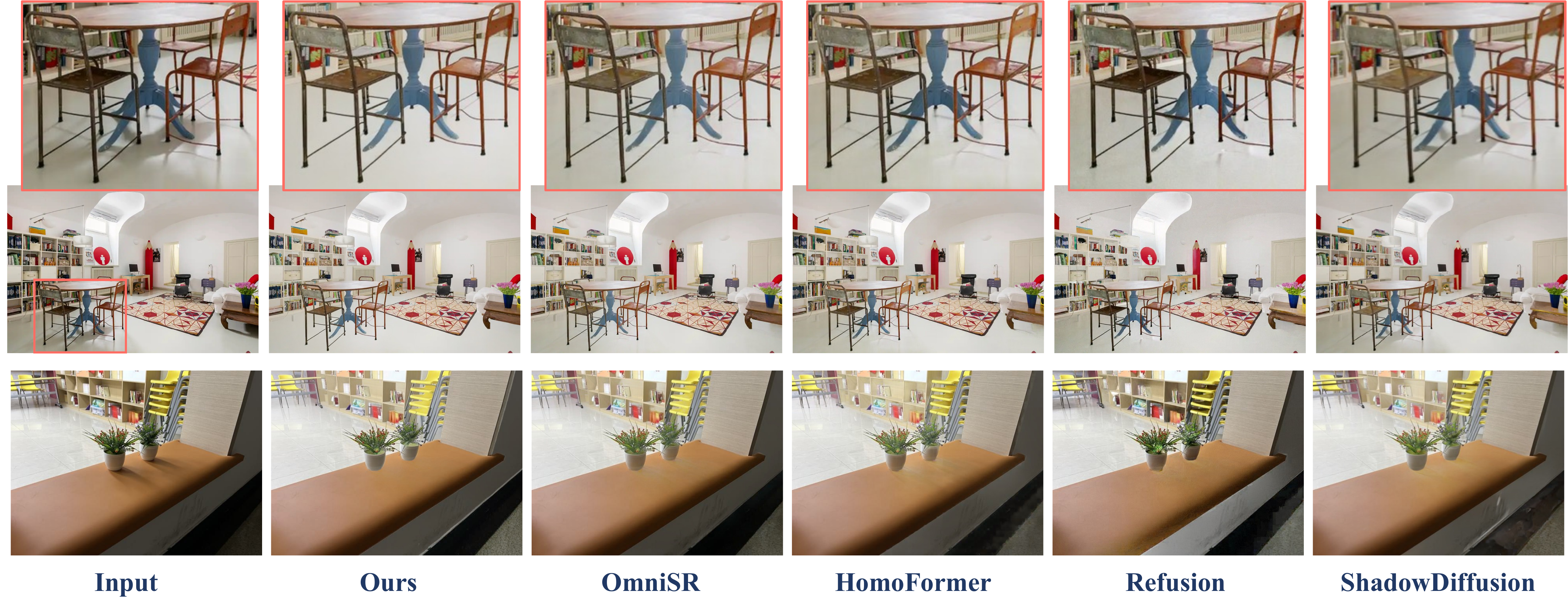}
\caption{\textbf{Comparisons with SOTA shadow removal methods in real scenes.} We compare our method to OmniSR~\cite{xu2024omnisr}, HomoFormer~\cite{xiao2024homoformer}, Refusion~\cite{luo2023refusion}, and ShadowDiffusion~\cite{guo2023shadowdiffusion}, all trained on the INS dataset~\cite{xu2024omnisr}. Results show that our approach achieves more comprehensive shadow removal, even in complex scenes.}
\label{fig:comparison_2} 
\end{figure*}

\paragraph{Datasets.} We conduct experiments on four shadow removal datasets: SRD~\cite{qu2017deshadownet}, ISTD+~\cite{wang2018stacked,le2019shadow}, WSRD+~\cite{vasluianu2023wsrd,vasluianu2024ntire}, and INS~\cite{xu2024omnisr}. The SRD dataset~\cite{qu2017deshadownet} includes 3,088 paired shadow and shadow-free images, divided into 2,680 for training and 408 for testing. The ISTD+ dataset~\cite{le2019shadow,wang2018stacked} provides 1,870 triplets of shadow images, shadow masks, and shadow-free images, with 1,330 used for training and 540 for testing. WSRD+\cite{vasluianu2023wsrd,vasluianu2024ntire} contains 1,000 pairs of shadow and non-shadow images captured under both spotlight and diffuse illumination. Since the test data is not open-sourced, we use its 100 evaluation images for testing. Lastly, the INS dataset~\cite{xu2024omnisr} comprises 30,000 synthetic training images under direct and indirect illumination, a synthetic test set of 2,000 images.


\paragraph{Evaluation Metrics.} 

Following previous methods \cite{fu2021auto,le2020shadow,guo2023shadowformer}, we evaluated images with a resolution of $256 \times 256$. We report results using the Peak Signal-to-Noise Ratio (PSNR) and the Structure Similarity Index Measure (SSIM)~\cite{wang2004image}, adopting the MATLAB evaluation codes as provided by Zhu et al. \cite{zhu2022efficient}. For the WSRD+~\cite{vasluianu2023wsrd} dataset, we used its evaluation data and the evaluation code provided by the NTIRE 2024 Image Shadow Removal Challenge~\cite{vasluianu2024ntire} for comparison.

\subsection{Comparisons}

We compare our method to twelve state-of-the-art shadow removal techniques, evaluating both quantitatively and qualitatively. Among these, eight methods require a shadow mask as input: AutoExposure~\cite{fu2021auto}, Zhu et al.~\cite{zhu2022efficient}, BMNet~\cite{zhu2022bijective}, ShadowFormer~\cite{guo2023shadowformer}, DMTN~\cite{liu2023decoupled}, ShadowDiffusion~\cite{guo2023shadowdiffusion}, and HomoFormer~\cite{xiao2024homoformer}. The remaining four methods—TBRNet~\cite{liu2023shadow}, Refusion~\cite{luo2023refusion}, Des3~\cite{jin2024des3}, and OmniSR~\cite{xu2024omnisr}—are mask-free, meaning they do not require a shadow mask as input. All comparisons use the original authors' implementations and hyperparameters, or the provided checkpoints or results, if available. For the INS dataset, we train these methods following the procedure in OmniSR~\cite{xu2024omnisr}.

For methods that are not mask-free, we use the shadow masks provided by the dataset as the default input. Since SRD~\cite{qu2017deshadownet} does not provide masks, we use the shadow masks generated by DHAN~\cite{cun2020towards}, which computes shadow matting from shadow/shadow-free pairs and generates binary masks through thresholding. In the case of the INS dataset, its shadow masks are represented as 0-1 probability maps rather than binary masks. Since it is a synthetic dataset with highly accurate masks, we found that using them as input results in methods producing outputs nearly identical to the ground truth (with PSNR values exceeding 50). To address this, following OmniSR~\cite{xu2024omnisr}, we use FDRNet~\cite{zhu2021mitigating} to detect shadow masks and input the detected masks instead.

As shown in Table~\ref{tab:comparison}, our method achieves the highest PSNR score on the INS dataset. On the ISTD+ dataset, our approach is the best mask-free method, performing only slightly below HomoFormer~\cite{xiao2024homoformer} and ShadowFormer~\cite{guo2023shadowformer}, which use ground-truth shadow masks as input, and outperforming most other methods that also rely on ground-truth shadow masks. As shown in the first two columns of Fig.~\ref{fig:comparison}, although without mask as input, our method is capable of thoroughly removing shadows, even with limited semantic context. We think it is because of leveraging the generative priors in our latent diffusion model. 

For the SRD dataset, our method also delivers competitive results compared to methods that utilize masks provided by DHAN~\cite{cun2020towards}. Among shadow-free methods, our performance is second only to DeS3. As illustrated in Fig.~\ref{fig:comparison}, our method produces visually appealing results with nearly complete shadow removal. We found that our method achieves better shadow removal quality than DeS3; however, in areas with highly detailed textures, such as sand and grass, the PSNR is lower than that of DeS3.


\begin{table}[t]
\small
\setlength\tabcolsep{2.5pt}
\center
\begin{tabular}{l|c c c c c c}
\hline
\multirow{2}{*}{Method} & \multicolumn{2}{c}{ISTD+→SRD} & \multicolumn{2}{c}{SRD→ISTD+} & \multicolumn{2}{c}{INS→WSRD+} \\ 
& {PSNR} & {SSIM} & {PSNR} & {SSIM} & {PSNR} & {SSIM} \\  
\hline
ShadowDiffusion & 24.26 & 0.890 & {29.48} & {0.947} & 19.40 & {0.825} \\ 
Refusion & 22.56 & 0.886 & 27.31 & 0.924 & 18.73 & 0.787  \\
DeS3 & 22.09 & 0.880 & --- & --- & --- & ---  \\
OmniSR & {25.86} & {0.917} & {30.19} & {0.951} & {20.03} & {0.825} \\
Ours & \textbf{26.23} & \textbf{0.923} & \textbf{30.23} & \textbf{0.953} & \textbf{20.11} & \textbf{0.828} \\
\hline
\end{tabular}
\caption{\textbf{Cross-dataset evaluation.} ISTD+→SRD means training on the ISTD+ dataset and tesing on the SRD dataset.}
\label{tab:generalizability}
\end{table}

On the INS dataset, our method achieves the highest PSNR and SSIM, as shown in Table~\ref{tab:comparison}. We also compare our method with others on real indoor scene images, all trained on the INS dataset. As illustrated in Fig.~\ref{fig:comparison_2}, our approach more effectively removes shadows from multiple light sources, such as those beneath the table and chair, as well as soft shadows behind the small planter and on the bookshelf in the background.

We further evaluated our method on high-resolution shadow removal using the WSRD+~\cite{vasluianu2023wsrd} dataset, with images at $1920 \times 1440$ resolution. The first stage requires no modification. We resize each image to $640 \times 480$ for the LDM fine-tuning. In the second stage, we introduce a minor adjustment by adding one patch-partition layer and one patch-merging layer~\cite{liu2021swin}. Additional details are provided in the supplementary materials. As shown in Table~\ref{tab:comparison}, our method outperforms others on the WSRD+ dataset, with more results available in the supplementary material. The relatively low PSNR scores for all methods on the WSRD+ dataset can be attributed to exposure differences between the input and ground-truth images.

\subsection{Cross-dataset evaluation.} 

To evaluate the generalizability of our approach, we conduct a cross-dataset evaluation. Specifically, we train our model on the training set of one dataset and test it on the testing set of a different dataset. Typically, training and testing data within the same dataset share similar shadow patterns, backgrounds, or lighting conditions. This cross-dataset evaluation presents a more challenging test for shadow removal methods and provides a more stringent measure of generalizability. We use $A$→$B$ to denote training on dataset $A$ and testing on dataset $B$. As shown in Table~\ref{tab:generalizability}, our method achieves the best results on ISTD+→SRD, SRD→ISTD, and INS→WSRD+. Here, we compare only with DeS3~\cite{jin2024des3} on ISTD+→SRD, as it provides checkpoints only for ISTD+ dataset. Its open-source code is a basic version that excludes certain components, including the VIT loss. For the INS→WSRD+ experiment, high-resolution images from the WSRD+ dataset are resized to $640 \times 480$ for inference and evaluation. Qualitative comparisons for this evaluation are provided in the supplementary material.



\begin{table}[t]
\small
\center
\begin{tabular}{c c c c}
\hline
\multirow{2}{*}{Dataset} & ISTD+ & SRD & INS \\
& PSNR$$/SSIM$$ & PSNR$$/SSIM$$ & PSNR$$/SSIM$$ \\
\hline
Full & \textbf{35.19/0.974} & \textbf{33.63/0.968} & \textbf{30.56/ 0.975} \\
W/o stage two & 29.95/0.843 & 28.18/0.868 & 28.78/0.924 \\ 
W/o DINO & 34.76/0.971 & 33.04/0.965 & 30.30/0.975 \\
\hline
\end{tabular}
\caption{\textbf{Ablation studies.} W/o DINO: without DINO in the detail injection model.}
\label{tab:ablation}
\end{table}

\subsection{Ablation study.}

To validate our model design, we conducted ablation studies on the two-stage pipeline and the addition of DINO features. The results are reported in Table~\ref{tab:ablation}. Stage two plays a critical role in our method, as removing it leads to a significant drop in PSNR and SSIM. This is due to stage one’s inability to effectively preserve high-frequency details. 
Additionally, incorporating DINO features into the detail injection model of stage two proves beneficial, enhancing PSNR and SSIM across all three datasets. For qualitative results related to the ablation study, please refer to the supplementary materials.



\section{Conclusion}

This paper introduces a novel detail-preserving latent diffusion pipeline that effectively eliminates complex shadows while preserving shadow-free details. The proposed method leverages the rich visual priors of a pre-trained Stable Diffusion model and propose a two-stage fine-tuning strategy to adapt the SD model for stable and efficient shadow removal. Extensive experiments demonstrate that our approach outperforms state-of-the-art shadow removal methods, particularly in terms of generalizability.

\paragraph{Limitation and future work.} Our method may still miss some shadows in complex scenes, such as the self-shadows on the flower pot in Fig.~\ref{fig:comparison_2}. In the future, we plan to incorporate unsupervised or self-supervised signals to further enhance the generalizability of our method.


{
    \small
    \bibliographystyle{ieeenat_fullname}
    \bibliography{main}
}


\end{document}